\documentclass{article}
\usepackage{ijcai24}

\usepackage{times}
\usepackage{soul}
\usepackage{url}
\usepackage[hidelinks]{hyperref}
\usepackage[utf8]{inputenc}
\usepackage[small]{caption}
\usepackage{graphicx}
\usepackage{amsmath}
\usepackage{amsthm}
\usepackage{booktabs}
\usepackage{algorithm}
\usepackage{algorithmic}
\usepackage[switch]{lineno}

\usepackage{multirow}

\usepackage{subcaption}

\title{Towards a Novel Perspective on Adversarial Examples Driven by Frequency}

\author{
Zhun Zhang$^1$
\and
Yi Zeng$^1$\and
Qihe Liu$^{1,*}$\And
Shijie Zhou$^1$
\affiliations
$^1$The School of Information and Software Engineering, University of Electronic Science and Technology of China, Chengdu 610054\\
\emails
\{zhunzhang, 202122090515\}@std.uestc.edu.cn,
\{qiheliu, sjzhou\}@uestc.edu.cn,
}

\begin{document}

\maketitle

\begin{abstract}

Enhancing our understanding of adversarial examples is crucial for the secure application of machine learning models in real-world scenarios. A prevalent method for analyzing adversarial examples is through a frequency-based approach. However, existing research indicates that attacks designed to exploit low-frequency or high-frequency information can enhance attack performance, leading to an unclear relationship between adversarial perturbations and different frequency components. In this paper, we seek to demystify this relationship by exploring the characteristics of adversarial perturbations within the frequency domain. We employ wavelet packet decomposition for detailed frequency analysis of adversarial examples and conduct statistical examinations across various frequency bands. Intriguingly, our findings indicate that significant adversarial perturbations are present within the high-frequency components of low-frequency bands. Drawing on this insight, we propose a black-box adversarial attack algorithm based on combining different frequency bands. Experiments conducted on multiple datasets and models demonstrate that combining low-frequency bands and high-frequency components of low-frequency bands can significantly enhance attack efficiency. The average attack success rate reaches 99\%, surpassing attacks that utilize a single frequency segment. Additionally, we introduce the normalized disturbance visibility index as a solution to the limitations of $L_2$ norm in assessing continuous and discrete perturbations\footnote{Our code will be released soon.}.
\end{abstract}

\section{Introduction}

\label{sec:intro}
With the advent of deep learning, neural network models have demonstrated groundbreaking performance across a variety of computer vision tasks \cite{aceto2021distiller,menghani2023efficient}. However, the emergence of adversarial examples has exposed vulnerabilities in deep neural networks (DNNs) \cite{goodfellow2014explaining,van2023anti}. These maliciously crafted adversarial examples are designed to induce misclassification in a target model by introducing human-imperceptible perturbations to clean examples. Such vulnerabilities impede the deployment of DNNs in security-sensitive domains.

Enhancing our comprehension of adversarial examples is pivotal for the development of trustworthy AI. A common method for examining adversarial examples is the signal-processing perspective of frequencies \cite{yin2019fourier,wang2020toward,lorenz2021detecting,deng2022frequency,qian2023lea2}. Typically, low-frequency components of images hold primary information, such as smooth backgrounds or uniform color areas, while high-frequency is associated with rapid changes in texture, edges, or noise.

However, adversarial examples in frequency domains remain a subject of debate \cite{han2023interpreting}. Goodfellow et al.\cite{goodfellow2014explaining} highlight the sensitivity of DNN activations to the high-frequency elements within images. Subsequent studies \cite{wang2020high,yin2019fourier} have underscored the impact of high-frequency signals on DNN predictions, while other research \cite{guo2018low,sharma2019effectiveness} posits that low-frequency perturbations can also effectively compromise models. Furthermore, Geirhos et al.\cite{geirhos2018imagenet} discover that DNNs trained on ImageNet are highly biased towards the texture (high-frequency) and shape (low-frequency) aspects of objects. Recent research reveals that adversarial examples are neither in high-frequency nor low-frequency components \cite{maiya2021frequency,qian2023lea2}. Consequently, the relationship between adversarial perturbations and frequency components remains uncertain, inviting further exploration in the frequency domain.


In this paper, we explore the relationship between adversarial perturbations and frequency components in the frequency domain. We utilize wavelet packet decomposition (WPD) to analyse adversarial perturbations, which provide a more nuanced analysis of local frequency characteristics than discrete cosine transform (DCT) or traditional wavelet analysis. Our analysis, supported by statistical data across different frequency bands, reveals a counterintuitive pattern: significant adversarial perturbations are predominantly present in the high-frequency components of low-frequency bands. Based on these findings, we propose a black-box adversarial attack algorithm that can add perturbations to any combinations of frequency bands separately. Experiments on three mainstream datasets and three models show that combining low-frequency bands and high-frequency components of low-frequency bands can effectively attack deep learning models, surpassing attacks that utilize a single frequency band. Moreover, we find there are limitations of $L_2$ norm in evaluating continuous and discrete perturbations when assessing the imperceptibility of adversarial examples. To overcome this, we introduce the normalized disturbance visibility (NDV) index, which normalizes the $L_2$ norm for a more accurate evaluation that aligns more closely with human perception.

The main contributions can be summarized as follows:

\begin{itemize}
	\item {Our research into the relationship between adversarial perturbations and frequency components in the frequency domain has uncovered a counterintuitive phenomenon: significant adversarial perturbations predominantly reside within the high-frequency parts of low-frequency bands. This finding has the potential to deepen our understanding of adversarial examples and to guide the design of adversarial attack and defense strategies.}
	
	
	\item{We propose a novel black-box adversarial attack algorithm leveraging frequency decomposition. Experiment results on three datasets (CIFAR-10, CIFAR-100 and ImageNet) and three models (ResNet-50, DenseNet-121 and InceptionV3) show that the strategy of combining low-frequency bands and high-frequency components of low-frequency significantly improves attack efficiency, with an average attack success rate of 99\%, outperforming approaches that utilize a single-frequency band.}
	
	\item{We introduce the NDV to address the issue of $L_2$ norm in assessing continuous and discrete perturbations, which provides a more comprehensive measure that aligns closely with human visual perception.}
	
	
	
	
\end{itemize}

\section{Related Works}
\label{sec:Related}
\subsection{Adversarial Example}
The notion of adversarial examples is initially introduced by Szegedy et al. \cite{goodfellow2014explaining}. They leverage gradient information to ingeniously engineer minor modifications in images. These alterations, generally imperceptible to the human eye, are sufficient to mislead classification algorithms into incorrect predictions. According to knowledge of the target model, adversarial attacks can be classified as white-box attacks or black-box attacks.
In white-box attack scenarios, attackers have access to gradients of models, enabling them to generate highly precise perturbations \cite{athalye2018synthesizing,carlini2017towards,kurakin2018adversarial,madry2017towards}.
Black-box attacks are broadly categorized into query-based and transfer-based types. Query-based attacks leverage feedback from model queries to craft adversarial examples \cite{chen2020hopskipjumpattack,cheng2018query,tu2019autozoom,guo2019simple,andriushchenko2020square}. Brendel et al. \cite{brendel2017decision} pioneered the decision-based black-box attack Cheng et al. \cite{chen2017zoo} introducing a zero-order optimization technique through gradient estimation. Conversely, transfer-based attacks involve creating adversarial examples on substitute models and transferring them to the target model \cite{cheng2019improving,huang2019black,huang2019enhancing,shi2019curls}.  However, compared to query-based attacks, transfer-based attacks are often constrained by challenges in building effective substitute models and their relatively lower rates of attack success.
\subsection{Frequency for Adversarial Example.}
Numerous studies \cite{tsuzuku2019structural,guo2018low,sharma2019effectiveness,lorenz2021detecting,wang2020toward} have analyzed DNNs from a frequency domain perspective. Tsuzuku \& Sato \cite{tsuzuku2019structural} pioneer the frequency framework by investigating the sensitivity of DNNs to different Fourier bases. Building on this, Guo et al. \cite{guo2018low} design pioneering adversarial attacks targeting low-frequency components. This approach is corroborated by Sharma et al. \cite{sharma2019effectiveness}, who demonstrate the efficacy of such attacks against models fortified with adversarial defenses. Concurrently, efforts by Lorenz et al. \cite{lorenz2021detecting} and Wang et al. \cite{wang2020toward} have propelled forward the detection of adversarial examples, utilizing frequency domain strategies during model training.

Additionally, some studies \cite{wang2020high,wang2020towards} have demonstrated that DNNs are more vulnerable to high-frequency components, resulting in models with low robustness. It is also the principal rationale for preprocessing-based defence methods. Confusingly, these findings are contradictory to the idea of low-frequency adversarial attacks\cite{ilyas2019adversarial,sharma2019effectiveness}. To explain the performance of adversarial examples in frequency domains, Maiya et al. \cite{maiya2021frequency} argue that adversarial examples are neither high-frequency nor low-frequency and are only relevant to the dataset. In addition, Abello et al. \cite{abello2021dissecting} and Caro et al. \cite{caro2020local} argue that the frequency of images is an intrinsic characteristic of the robustness of models. Recent studies \cite{qian2023lea2} have shown that attacks targeting both low-frequency and high-frequency information can enhance attack performance, indicating that the relationship between learning behavior and different frequency information remains to be explored.


\section{Frequency Analysis of Adversarial Perturbations}
\label{sec:Meth}

In this section, we introduce the steps to analyze adversarial perturbation in the frequency domain, including 1) adversarial perturbation generation, 2) frequency decomposition, and 3) frequency analysis.


\subsection{Adversarial Perturbation Generation}
\label{fc}
To examine the characteristics of adversarial perturbations, we first generate these disruptive inputs. FGSM \cite{goodfellow2014explaining} and PGD \cite{madry2017towards} are prominent techniques for creating adversarial examples, commonly used to assess the robustness of deep learning models. Consequently, we employ these two methods for generating adversarial examples. To be specific, given some images $X=\{x_1,x_2,..., x_n\}$, whose corresponding labels $Y =\{y_1, y_2,..., y_n\}$. The adversarial example generation formula for FGSM is as Eq.\ref{eq0}:

\begin{equation}
x_{adv} = x + \epsilon \cdot sign(\nabla_x J(\theta, x, y))
\label{eq0}
\end{equation}

where, $\epsilon$ is a parameter that controls the magnitude of the perturbation. $sign(\nabla_x J(\theta, x, y))$ represents the sign of the gradient of the loss function with respect to the input image, which determines the direction of the perturbation. PGD generates an adversarial example by performing the iterative update with Eq.\ref{eq1}:
\begin{equation}
x^{t+1}_{adv}=P_{x,\epsilon}(x^t_{adv}+\alpha^{t}\cdot sign(\nabla_{x}J(x_t,y)))
\label{eq1}
\end{equation}

where $\alpha^{t}$ is the step size at iteration $t$, $x^t_{adv}$ is adversarial example at iteration $t$ and $P_{x,\epsilon}(\cdot)$ clips the input to the $\epsilon-$ball of $x$. $sign(\nabla_{x}J(x_t,y))$ is the gradient at iteration $t$.

\begin{figure}[t]
	\centering
	\includegraphics[width=1\linewidth]{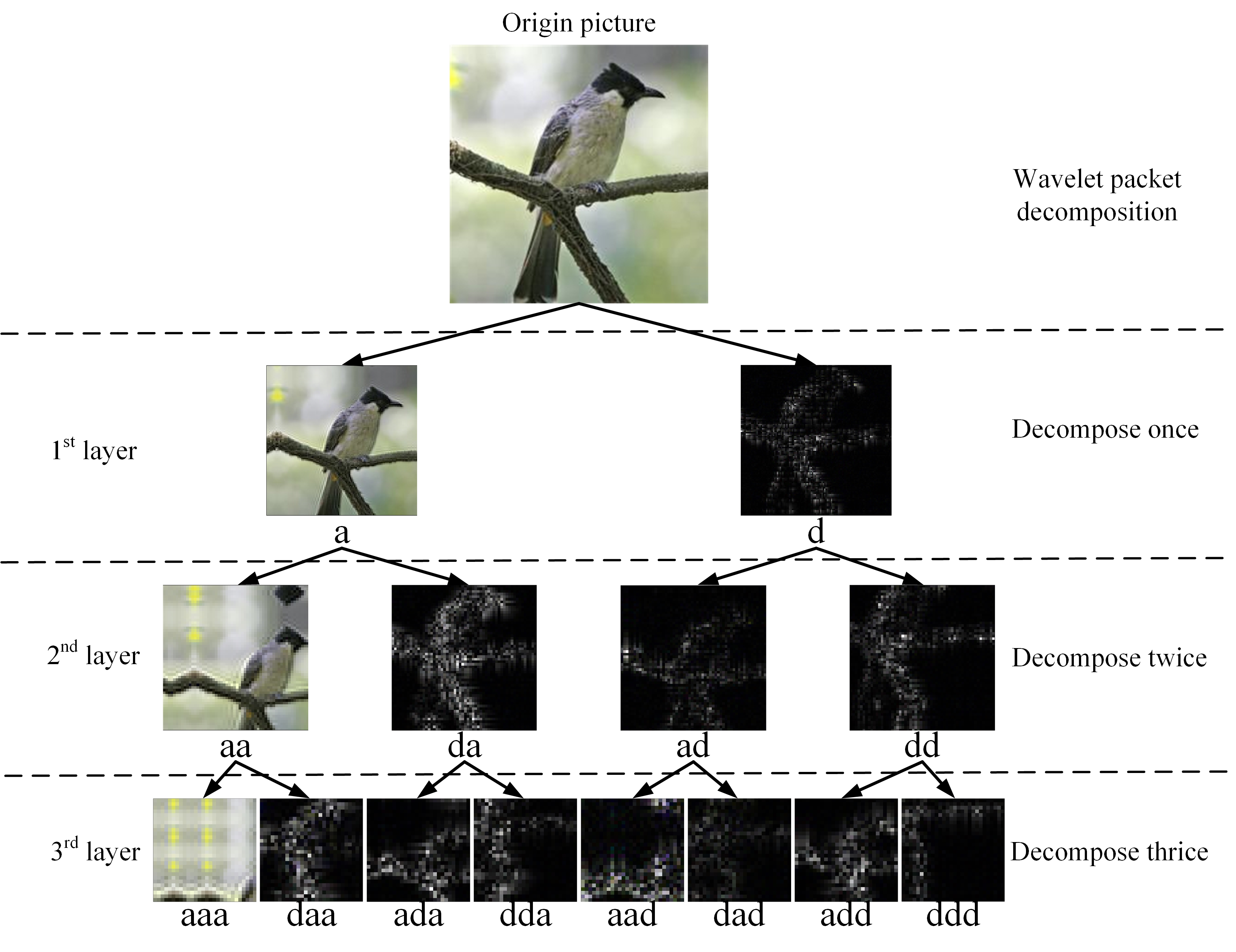}
	\caption{After an initial decomposition, the original image is split into two components: ${a}$, representing the low-frequency band, and ${d}$, representing the high-frequency band at the $1^{st}$ layer.  Subsequent decomposition of both ${a}$ and ${d}$ results in four distinct frequency bands at the $2^{nd}$ layer. This process continues, with the bands from the $2^{nd}$ layer being further decomposed to yield eight frequency bands at the $3^{rd}$ layer. Among these, the $\{aaa\}$ band preserves the core attributes of the original image, such as color, brightness, and general structure. Conversely, the other seven bands are primarily linked to edge definition.}
	\label{fig6}
	\vspace{-0.1in}
\end{figure}

\subsection{Frequency Decomposition}
Wavelet packet decomposition (WPD) employs multi-level data decomposition to extract detailed and approximate information from signals. This method iteratively decomposes all frequency components, enabling a comprehensive frequency stratification. This contrasts traditional wavelet analysis, which primarily decomposes only the low-frequency parts. Through WPD, an image is separated into a series of frequency bands, each capturing distinct components of the image. The low-frequency band typically encapsulates fundamental image characteristics, such as overall luminosity, color, and structural composition. In contrast, high-frequency bands contain secondary image information like edges and textures \cite{prasad2020wavelet}.


We apply WPD for frequency decomposition. The wavelet packet $\{\mu_{n}(t)\}_{n\in Z}$ comprises a set of functions that include the scaling function $\mu_{0}(t)$ and the wavelet function $\mu_{1}(t)$, as depicted in Eq.\ref{eqwt}:

\begin{equation}
\begin{cases}
\mu_{2n}(t)=\sum_{k\in Z}p_{k}\mu_{n}(2t-k)\\
\mu_{2n+1}(t)=\sum_{k\in Z}q_{k}\mu_{n}(2t-k)
\end{cases}
\label{eqwt}
\end{equation}

where $p_{k}$ is low-pass filter, $q_{k}$ is high-pass filter. The decomposition process of WPD can be expressed as Eq.\ref{eqwtd}:

\begin{equation}
\begin{cases}
f^{2n}_{jk}=\frac{1}{\sqrt{2}}\sum_{l\in Z}p_{l-2k}{f^{n}_{j+l,l}}\\
f^{2n+1}_{jk}=\frac{1}{\sqrt{2}}\sum_{l\in Z}q_{l-2k}{f^{n}_{j+l,l}}
\end{cases}
\label{eqwtd}
\end{equation}

where $f^{2n}_{jk}$ and $f^{2n+1}_{jk}$ are wavelet packet coefficients obtained by the decomposition of ${f^{n}_{j+l,l}}$. So we decompose $X$, $X_{adv}$ into frequency domains by Eq.\ref{eqwtd}, and we get different wavelet packet coefficients $F=\{f_1,f_2,...,f_n\}$ and  $F_{adv}=\{{f_1^\prime},{f_2^\prime},...,{f_n^\prime}\}$ respectively. The wavelet packet coefficient determines the frequency band, so we think of them as the frequency band.

For ease of description, high-frequency components are denoted as $d$, and low-frequency components as $a$. WPD follows a binary decomposition process, resulting in a tree structure as illustrated in Figure \ref{fig6}. After $n$ decompositions, $2^n$ frequency bands are obtained. These decomposed frequency bands can be reassembled using Eq.\ref{eqre}:

\begin{equation}
f^{n}_{j-1,k}=\frac{1}{\sqrt{2}}\sum_{l\in Z}(p_{k-2l}{f^{2n}_{jl}}+q_{k-2l}f_{jl}^{2n+1})
\label{eqre}
\end{equation}

\subsection{Frequency Analysis}
\label{sec3.3}
We employ WPD to dissect both normal images and their corresponding adversarial examples into a series of frequency bands. This approach allows us to compute the cosine similarity between adversarial and clean samples within each band, thereby analyzing the spread of adversarial perturbations across different frequency segments. For an image $X$ and its adversarial counterpart $X_{adv}$, we apply WPD to decompose $X$ into frequency bands $F=\{f_1,f_2,...,f_n\}$ and $X_{adv}$ into $F_{adv}=\{f_1^\prime, f_2^\prime,..., f_n^\prime\}$. Subsequently, we calculate their cosine similarity $cos(\theta)$ as per Eq.\ref{eqcos}.

\begin{equation}
cos(\theta)=\frac{|f_i\cdot f_i^\prime|}{{||f_i||}_2{||f_i^\prime||}_2}, i=1,2,...,n
\label{eqcos}
\end{equation}

\begin{figure}[h]
	\centering
	\includegraphics[width=0.9\linewidth]{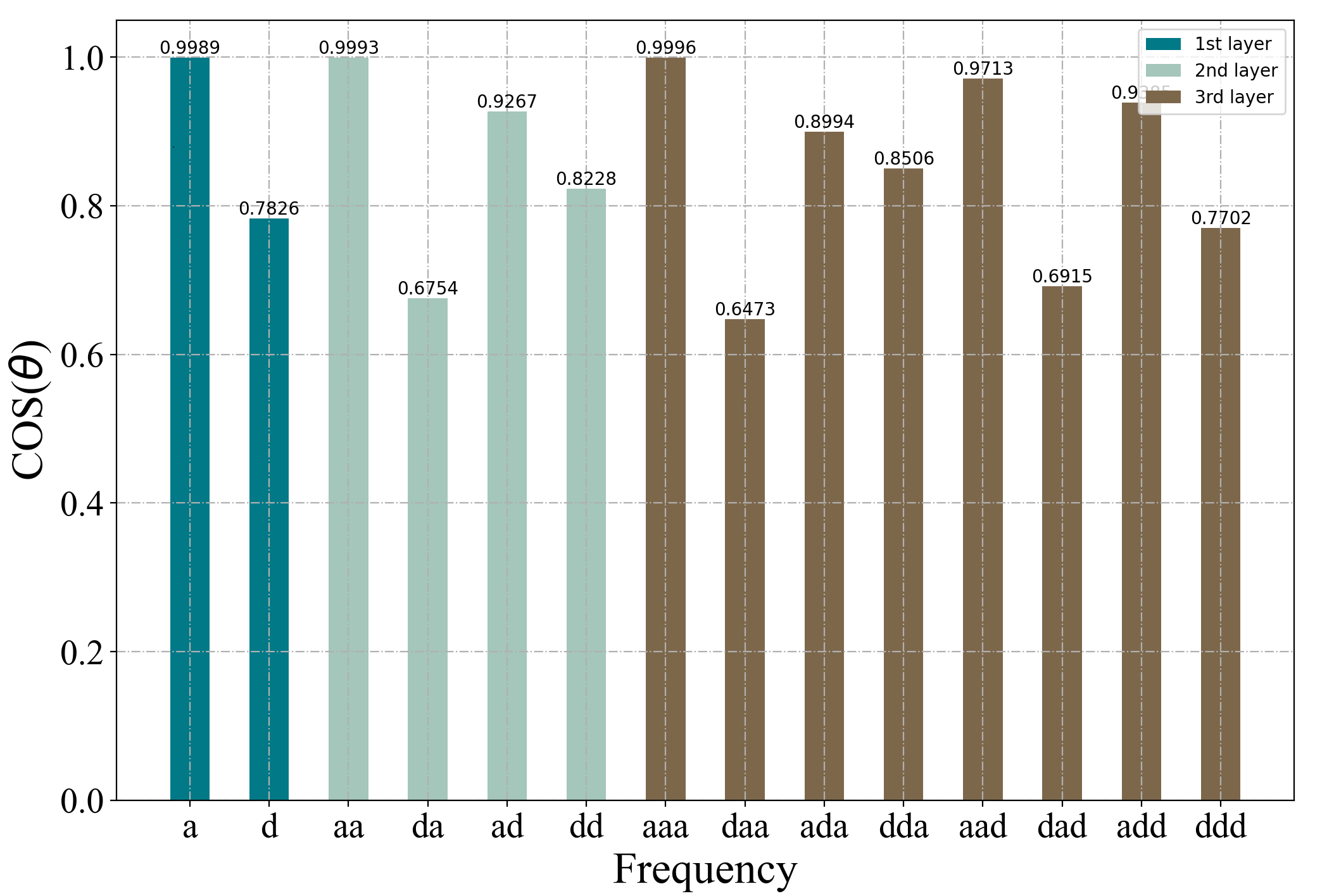}
	\caption{This analysis compares the average cosine similarity between adversarial and pure examples across different frequency bands. The first decomposition yields the $\{a, d\}$ bands, and subsequent decompositions of $\{a\}$ and $\{d\}$ result in the $\{aa, da\}$ and $\{ad, dd\}$ bands, and so forth.}
	\label{fig3}
	\vspace{-0.1in}
\end{figure}

The experimental results are depicted in Figure \ref{fig3}, demonstrating that lower similarity corresponds to greater perturbations. Initially, after the first decomposition ($1^{st}$ layer), adversarial perturbations predominantly appear in the high-frequency component $\{d\}$, aligning with previous research that identifies adversarial perturbations as high-frequency components \cite{wang2020high,wang2020towards}. Interestingly, after the second decomposition ($2^{nd}$ layer), these perturbations concentrate more in the $\{da\}$ band, with further decrease in similarity observed in the $\{daa\}$ band after the third decomposition ($3^{rd}$ layer). The $\{dad\}$ band which decomposes from the low-frequency band ${ad}$ originating from the high band $\{d\}$, also exhibits reduced similarity. This pattern suggests that adversarial perturbations are prevalent in higher-frequency components within low-frequency bands. Furthermore, at the third decomposition level, the band with the third-lowest similarity is the highest frequency band $\{ddd\}$, indicating that adversarial perturbations, although inherently high-frequency, depend on the low-frequency information from their preceding levels. For more details, see Section \ref{eapf}.

\begin{algorithm}[h]
	\caption{Adversarial Attack based on Frequency Decomposition}
	\label{alg1}
	\begin{algorithmic}[1]
		\REQUIRE Pure images $x$, Label $y$, Number $n$, Step length $\epsilon$   \\
		\ENSURE Adversarial examples $x_{adv}$\\
		\STATE  Initial $x_{adv}=x$, Decompose $x_{adv}$ into frequency bands $W$ with Eq.\ref{eqwtd};\
		\STATE  Select $W_s \subset W$;\
		\STATE Initializing a probability $Pro$, which is the probability of choosing $Q \in W_s$; 
		\STATE Confidence queried by the model $p=ph(y|x_{adv})$;\
		
		\WHILE {$p_y = \max_{y^\prime}p_{y^\prime}$}
		\FOR{$\alpha \in \{\epsilon,-\epsilon\}$}
		
		\STATE	According to $Pro$, pick $Q \in W_s$;\ 
		\STATE	Initialize null matrix $q$ of size $Q$;\
		\STATE  Randomly pick $n$ points in $q$ without replacement, and assigns selected points a value of 1;\
		\STATE	$Q=Q+\alpha q$;\
		\STATE Reconstruct the perturbed ${W}$ separately to get $x^\prime$ with Eq.\ref{eqre};\

		
		
		\IF{$ph(y|x^\prime)<p$}
		\STATE $p=ph(y|x^\prime)$;\
		\STATE $x_{adv}=x^\prime$;\
		\STATE Update $Pro$ according to $p$;\
		\ENDIF
		\ENDFOR
		
		\ENDWHILE
		
	\end{algorithmic}
	
\end{algorithm}

\section{Black-box Adversarial Attack Based on Frequency Decomposition}
\begin{figure*}[t]
	\centering
	\includegraphics[width=0.9\linewidth]{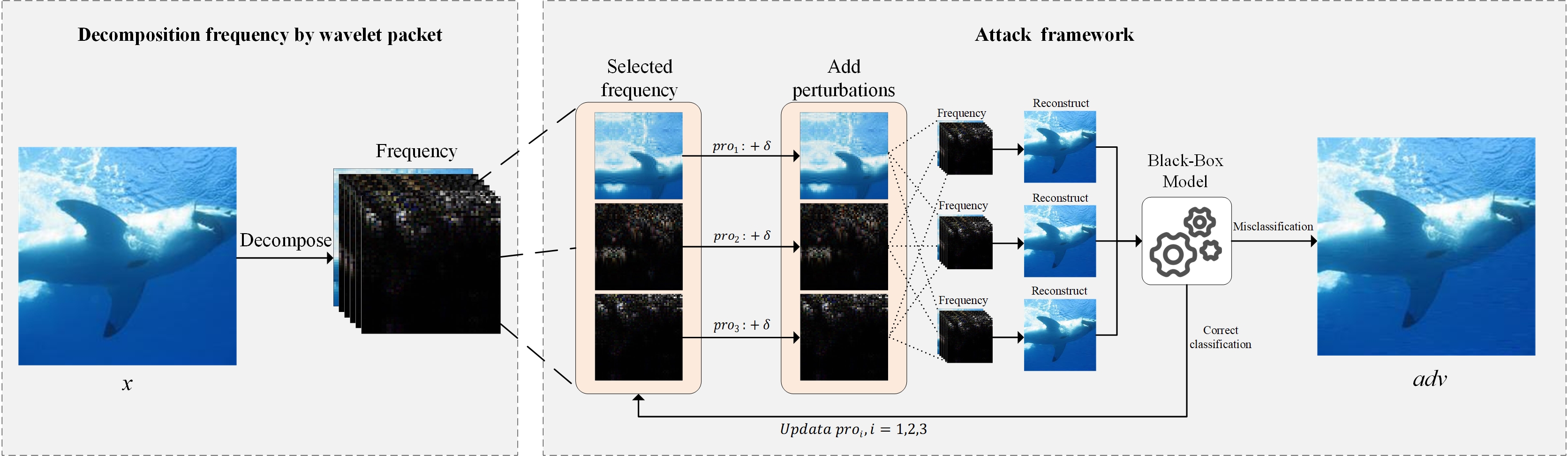}
	\caption{The proposed attack framework. This method is divided into two parts: (a) decomposing the digital image into multiple frequency bands using wavelet packet decomposition, and (b) implementing adversarial attacks based on frequency bands selection and perturbation addition strategies.}
	\label{fig2}
	\vspace{-0.1in}
\end{figure*}

In this section, we introduce a black-box adversarial attack methodology grounded in frequency decomposition. The framework of our attack is depicted in Figure \ref{fig2}, accompanied by the corresponding pseudo-code in Algorithm \ref{alg1}. The core components include frequency band selection and attack implementation.

\subsection{Frequency Selection}


As discussed in Section \ref{sec3.3}, our experimental findings indicate that adversarial perturbations predominantly appear in specific high-frequency bands originating from low-frequency bands. Particularly at the $3^{rd}$ layer, the $\{daa\}$ and $\{dad\}$ bands, decompose from $\{aa\}$ and $\{ad\}$ respectively, exhibit lower similarity. This indicates a higher concentration of adversarial perturbations in these bands. The lowest frequency band $\{aaa\}$ encompasses crucial image information like color, brightness, and structural luminance, contrasting with the other seven bands, which are mainly associated with edges and textures. This difference implies that the $\{aaa\}$ band holds more substantial information, as illustrated in Figure \ref{fig6}, where areas with lower information content are in black (value of 0). Consequently, we combine the low-frequency band $\{aaa\}$ and the high-frequency components of low-frequency bands $\{daa, dad\}$ as the main search area. The ablation experiments for these three frequency bands are detailed in Section \ref{ab}.

\subsection{Attack Implementation}
Inspired by the work of Guo et al \cite{guo2019simple}, we propose a black-box adversarial attack. This approach includes orthogonal direction vector selection, iterative approach and perturbation limitations.

\textbf{Orthogonal direction vector selection.} To effectively reduce the number of searches, verifying that all vectors in frequency bands $W_s$ are orthonormal is essential. Decompose space $L^2(R)$ into wavelet subspace $W_j$, such as $L^2(R)=\oplus_{j\in Z} W_j$. We further orthogonal decompose the subspace $W_j,j>1$ into $W_j=\oplus_{n\in Z} U^n_j$. According to wavelet theory, wavelet packet decompositions satisfy Eq.\ref{eqort}. Therefore, after three times decomposition, we yield $\{aaa, daa, ada, dda, aad, dad, add, ddd \}$, which are orthogonal to each other. So the frequency band $Q \in W_s$ are mutually orthogonal. The $q$, which we choose from $Q$ without repeat, is also orthogonal, where $q$ is the orthogonal direction vector specified in this paper.

\begin{equation}
\begin{cases}
U_j^{2n}\bot U_j^{2n+1},\\
U^{2n}_{j+1}= U_j^{2n}\oplus U_j^{2n+1},\\
j=0,1,2,...
\end{cases}
\label{eqort}
\end{equation}

\textbf{Iterative approach.} Pick band $Q$ from $W_s$ according to the initialization probability $Pro$. Then, initialize a matrix $q$ of size $Q$ with the value $1$ for $n$ random points and $0$ for the rest. No point in $q$ will be selected twice. Therefore, each $q \in Q$ is an orthogonal vector we picked. We add perturbations to the frequency band $Q$ through $\alpha q$. Since $Q$ belongs to $W_s$ and $W_s$ belongs to $W$, our perturbations are added to $W$. We can reconstruct the perturbed $W$ to image $x^\prime$ by Eq.\ref{eqre}. Input $x^\prime$ into the model for a query. These modifications may likely lower the $p=ph (y|x^\prime)$. And we get the $p$ in the current search. Then, we update the probability $Pro$ based on the $p$ and choose another orthogonal vector $q$ for the subsequent iteration. To maintain the query efficiency, we ensure that no two directions cancel each other out.

\textbf{Perturbation limitations.} We can control perturbations by adjusting the step size $\epsilon$ and $n$ (parameter $n$ determines the generation of $q$). In each iteration, $n$ points are added, subtracted, or discarded (if the output probability is not reduced in either direction). Suppose that $\alpha \in \{-\epsilon,0,\epsilon\}$ represents the symbol of the search direction chosen at step $T$. Therefore,  perturbations after $t$ steps in frequency domains satisfy Eq.\ref{eq6}. Our analysis highlights a critical trade-off: (1) for query-limited situations, raising $\epsilon$ and $n$ can increase the attack success rate while decreasing the number of inquiries. (2) Reduce $\epsilon$ and $n$ to produce imperceptible perturbations in perturbation-limited situations.


\begin{equation}
\delta_T = \sum_{t=1}^{T} \alpha_tq_t
\label{eq6}
\end{equation}

\begin{figure}[h]
	\centering
	\includegraphics[width=0.8\linewidth]{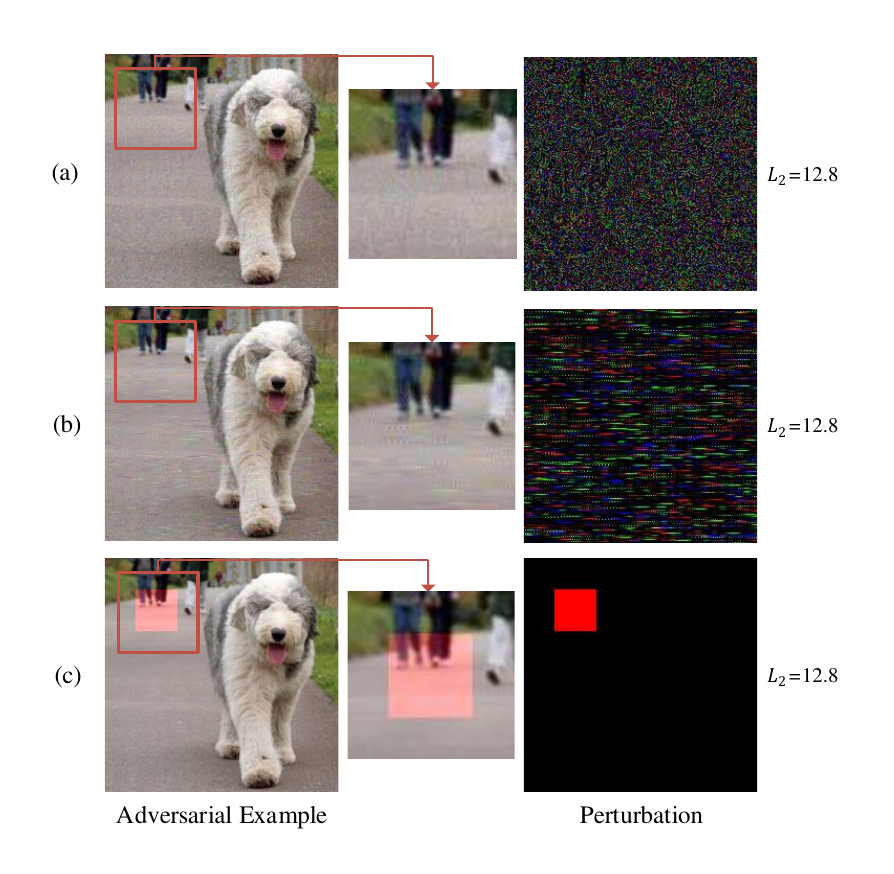}
	
	\caption{We compare some adversarial examples and their perturbations, all normalized to the same $L_2$ distance. These include (a) PGD, (b) our attack, and (c)  Patch attack. For visualization purposes, the perturbations have been amplified by a factor of 10.}
	\label{fig1}
	\vspace{-0.1in}
\end{figure}

\begin{table*}[]
	\centering
	\scalebox{0.72}{
\begin{tabular}{lll|rr|rrrr|rrrrrrrr}
	\toprule
	\multirow{2}{*}{Datasets} & \multirow{2}{*}{Models} & \multicolumn{1}{c|}{\multirow{2}{*}{Attacks}} & \multicolumn{2}{c|}{$1^{st}$ layer}                          & \multicolumn{4}{c|}{$2^{nd}$ layer}                                                                              & \multicolumn{8}{c}{$3^{nd}$ layer}                                                                                                                                                                                         \\ \cline{4-17} 
	&                         & \multicolumn{1}{c|}{}                         & \multicolumn{1}{c}{a} & \multicolumn{1}{c|}{\textbf{d}} & \multicolumn{1}{c}{aa} & \multicolumn{1}{c}{\textbf{da}} & \multicolumn{1}{c}{ad} & \multicolumn{1}{c|}{dd} & \multicolumn{1}{c}{aaa} & \multicolumn{1}{c}{\textbf{daa}} & \multicolumn{1}{c}{ada} & \multicolumn{1}{c}{dda} & \multicolumn{1}{c}{aad} & \multicolumn{1}{c}{\textbf{dad}} & \multicolumn{1}{c}{add} & \multicolumn{1}{c}{ddd} \\ \toprule
	\multirow{8}{*}{CIRAR10}  & \multirow{2}{*}{Den}    & FGSM 
	& 0.9986 & \textbf{0.7915} & 0.9992  & \textbf{0.6879} & 0.9367  & 0.8249  & 0.9995  & \textbf{0.6778}  & 0.9133  & 0.8574 & 0.9763 & \textbf{0.6906}  & 0.9483  & 0.7659 \\
	&                         & PGD                                           & 0.9991                & \textbf{0.8563}                 & 0.9994                 & \textbf{0.7649}                 & 0.9608                 & 0.8837                  & 0.9996                  & \textbf{0.7525}                  & 0.9456                  & 0.9069                  & 0.9849                  & \textbf{0.7699}                  & 0.9680                  & 0.8394                  \\
	& \multirow{2}{*}{R50}    & FGSM                                          & 0.9986                & \textbf{0.7987}                 & 0.9991                 & \textbf{0.6984}                 & 0.9366                 & 0.8301                  & 0.9994                  & \textbf{0.6905}                  & 0.9139                  & 0.8608                  & 0.9756                  & \textbf{0.7002}                  & 0.9477                  & 0.7754                  \\
	&                         & PGD                                           & 0.9992                & \textbf{0.8646}                 & 0.9995                 & \textbf{0.7776}                 & 0.9621                 & 0.8904                  & 0.9996                  & \textbf{0.7645}                  & 0.9479                  & 0.9122                  & 0.9853                  & \textbf{0.7830}                  & 0.9687                  & 0.8488                  \\
	& \multirow{2}{*}{V3}     & FGSM                                          & 0.9990                & \textbf{0.7387}                 & 0.9995                 & \textbf{0.5610}                 & 0.9457                 & 0.8271                  & 0.9997                  & \textbf{0.4965}                  & 0.9206                  & 0.8641                  & 0.9831                  & \textbf{0.6076}                  & 0.9583                  & 0.7639                  \\
	&                         & PGD                                           & 0.9994                & \textbf{0.8528}                 & 0.9997                 & \textbf{0.7174}                 & 0.9694                 & 0.9043                  & 0.9998                  & \textbf{0.6714}                  & 0.9559                  & 0.9256                  & 0.9897                  & \textbf{0.7475}                  & 0.9758                  & 0.8627                  \\
	& \multirow{2}{*}{Vgg}    & FGSM                                          & 0.9989                & \textbf{0.7591}                 & 0.9994                 & \textbf{0.6161}                 & 0.9405                 & 0.8192                  & 0.9996                  & \textbf{0.5751}                  & 0.9139                  & 0.8535                  & 0.9803                  &\textbf{0.6426}                  & 0.9539                  & 0.7594                  \\
	&                         & PGD                                           & 0.9994                & \textbf{0.8472}                 & 0.9996                 & \textbf{0.7255}                 & 0.9656                 & 0.8915                  & 0.9998                  & \textbf{0.6868}                  & 0.9494                  & 0.9128                  & 0.9887                  & \textbf{0.7494}                 & 0.9733                  & 0.8498                  \\ \hline
	\multirow{8}{*}{CIRAR100} & \multirow{2}{*}{Den}    & FGSM                                          & 0.9985                & \textbf{0.7884}                 & 0.9990                 & \textbf{0.6955}                 & 0.9280                 & 0.8172                  & 0.9993                  & \textbf{0.6780}                  & 0.8994                  & 0.8432                  & 0.9733                  & \textbf{0.7031}                  & 0.9416                  & 0.7667                  \\
	&                         & PGD                                           & 0.9990                & \textbf{0.8504}                 & 0.9993                 & \textbf{0.7643}                 & 0.9547                 & 0.8766                  & 0.9995                  & \textbf{0.7471}                  & 0.9364                  & 0.8976                  & 0.9828                  & \textbf{0.7721}                 & 0.9629                  & 0.8342                  \\
	& \multirow{2}{*}{R50}    & FGSM                                          & 0.9984                & \textbf{0.8084}                 & 0.9990                 & \textbf{0.7261}                 & 0.9238                 & 0.8319                  & 0.9992                  & \textbf{0.7070}                  & 0.8937                  & 0.8509                  & 0.9727                  & \textbf{0.7350}                  & 0.9387                  & 0.7926                  \\
	&                         & PGD                                           & 0.9990                & \textbf{0.8601}                 & 0.9993                 & \textbf{0.7866}                 & 0.9518                 & 0.8806                  & 0.9995                  & \textbf{0.7706}                  & 0.9311                  & 0.8970                  & 0.9825                  & \textbf{0.7935}                  & 0.9616                  & 0.8467                  \\
	& \multirow{2}{*}{V3}     & FGSM                                          & 0.9988                & \textbf{0.7468}                 & 0.9993                 & \textbf{0.5843}                 & 0.9445                 & 0.8305                  & 0.9996                  & \textbf{0.5255}                  & 0.9220                  & 0.8671                  & 0.9795                  & \textbf{0.6278}                  & 0.9549                  & 0.7675                  \\
	&                         & PGD                                           & 0.9992                & \textbf{0.8459}                 & 0.9995                 & \textbf{0.7123}                 & 0.9632                 & 0.8948                  & 0.9997                  & \textbf{0.6658}                  & 0.9500                  & 0.9171                  & 0.9856                  & \textbf{0.7410}                  & 0.9690                  & 0.8498                  \\
	& \multirow{2}{*}{Vgg}    & FGSM                                          & 0.9988                & \textbf{0.7532}                 & 0.9993                 & \textbf{0.6120}                 & 0.9396                 & 0.8146                  & 0.9996                  & \textbf{0.5681}                  & 0.9142                  & 0.8503                  & 0.9782                  & \textbf{0.6394}                  & 0.9515                  & 0.7517                  \\
	&                         & PGD                                           & 0.9994                & \textbf{0.8506}                 & 0.9996                 & \textbf{0.7321}                 & 0.9671                 & 0.8932                  & 0.9998                  & \textbf{0.6940}                  & 0.9528                  & 0.9154                  & 0.9884                  & \textbf{0.7549}                  & 0.9734                  & 0.8492                  \\ \hline
	\multirow{8}{*}{ImageNet} & \multirow{2}{*}{Den}    & FGSM                                          & 0.9983                & \textbf{0.6885}                 & 0.9989                 & \textbf{0.5980}                 & 0.8451                 & 0.7216                  & 0.9993                  & \textbf{0.5766}                  & 0.8042                  & 0.7549                  & 0.9272                  & \textbf{0.6079}                  & 0.8603                  & 0.6603                  \\
	&                         & PGD                                           & 0.9993                & \textbf{0.7657}                 & 0.9996                 & \textbf{0.6694}                 & 0.9215                 & 0.8026                  & 0.9997                  & \textbf{0.6426}                  & 0.8869                  & 0.8347                  & 0.9721                  & \textbf{0.6836}                  & 0.9351                  & 0.7421                  \\
	& \multirow{2}{*}{R50}    & FGSM                                          & 0.9985                & \textbf{0.6625}                 & 0.9991                 & \textbf{0.5821}                 & 0.8452                 & 0.6907                  & 0.9994                  & \textbf{0.5690}                  & 0.7918                  & 0.7251                  & 0.9407                  & \textbf{0.5856}                  & 0.8691                  & 0.6290                  \\
	&                         & PGD                                           & 0.9994                & \textbf{0.7655}                 & 0.9996                 & \textbf{0.6750}                 & 0.9239                 & 0.7995                  & 0.9997                  & \textbf{0.6525}                  & 0.8884                  & 0.8317                  & 0.9748                  & \textbf{0.6861}                  & 0.9387                  & 0.7394                  \\
	& \multirow{2}{*}{V3}     & FGSM                                          & 0.9982                & \textbf{0.6927}                 & 0.9988                 & \textbf{0.6101}                 & 0.8314                 & 0.7197                  & 0.9993                  & \textbf{0.5874}                  & 0.7896                  & 0.7441                  & 0.9142                  & \textbf{0.6188}                  & 0.8447                  & 0.6644                  \\
	&                         & PGD                                           & 0.9993                & \textbf{0.7665}                 & 0.9996                 & \textbf{0.6699}                 & 0.9119                 & 0.8016                  & 0.9997                  & \textbf{0.6407}                  & 0.8769                  & 0.8282                  & 0.9662                  & \textbf{0.6837}                  & 0.9238                  & 0.7429                  \\
	& \multirow{2}{*}{Vgg}    & FGSM                                          & 0.9983                & \textbf{0.6685}                 & 0.9989                 & \textbf{0.5788}                 & 0.8486                 & 0.7022                  & 0.9993                  & \textbf{0.5551}                  & 0.8016                  & 0.7342                  & 0.9368                  & \textbf{0.5916}                  & 0.8682                  & 0.6430                  \\
	&                         & PGD                                           & 0.9993                & \textbf{0.7608}                 & 0.9996                 & \textbf{0.6661}                 & 0.9222                 & 0.7994                  & 0.9997                  & \textbf{0.6392}                  & 0.8869                  & 0.8305                  & 0.9732                  & \textbf{0.6817}                 & 0.9364                  & 0.7412                  \\ \bottomrule
	\multicolumn{3}{c|}{AVE}                                                                            & 0.9989                & \textbf{0.7826}                 & 0.9993                 & \textbf{0.6755}                 & 0.9267                 & 0.8228                  & 0.9996                  & \textbf{0.6473}                  & 0.8994                  & 0.8506                  & 0.9713                  & \textbf{0.6915}                  & 0.9385                  & 0.7702                  \\ \bottomrule
\end{tabular}
}
\caption{The comparison of the cosine similarity between adversarial and pure examples across various frequency bands. We highlight the frequency bands with the lowest cosine similarity in each layer.}
\label{tab22}
\vspace{-0.1in}
\end{table*}

\section{Normalized Disturbance Visibility Index}

The $L_2$ norm is widely used to measure the imperceptibility by quantifying the perturbation magnitude between original and adversarial examples. It gauges the overall effect of the adversarial perturbation on the original input, helping to ensure that the resulting adversarial samples remain either imperceptible or minimally perceptible to human observers. However, our findings suggest that the $L_2$ norm might fall short in accurately evaluating adversarial examples that comprise continuous and discrete perturbations, as shown in Figure \ref{fig1}. Even with an equivalent $L_2$ norm, perturbations from patch attacks (see Figure \ref{fig1}(c)) tend to be more discernible to the human eye.

\textbf{Normalized Disturbance Visibility Index.}  To align more closely with human perceptual observations, we refine the $L_2$ norm into a more sophisticated metric, the Normalized Disturbance Visibility (NDV) Index. For a pure sample $x$ and its corresponding adversarial example $x_{adv}$, the NDV is computed as outlined in Eq.\ref{l2d}. This index modifies the $L_2$ norm by dividing it by the number of points affected by the perturbation. To circumvent issues of non-differentiability, we introduce a small constant $\epsilon$ in the denominator. Moreover, for standardizing the scale measurements, we multiply the final value by a constant $C$ (defaulted to 1000). The NDV values, akin to the $L_2$ norm, convey the magnitude of perturbations: a larger value indicates more significant perturbations, while a smaller value denotes less pronounced perturbations.

\begin{equation}
NDV=C\frac{{||x-x_{adv}||}_2}{{||x-x_{adv}||}_0+\epsilon}
\label{l2d}
\end{equation}

\section{Experiments}
\label{Ex}
\subsection{Datasets and Models}
Following the previous study \cite{guo2019simple}, we randomly pick 1000 images from validation sets of CIFAR-10 \cite{krizhevsky2009learning}, CIFAR-100 \cite{krizhevsky2009learning} and ImageNet-1K \cite{russakovsky2015imagenet}, respectively, as test dataset. These samples are initially correctly classified to avoid artificially boosting the success rate. In CIFAR-10 and CIFAR-100, we are standardized training the models of ResNet-50, VGG16 and InceptionV3 following the steps outlined in \cite{he2016deep,huang2017densely,szegedy2016rethinking}. On ImageNet-1K, we choose pre-trained models of ResNet-50, VGG16, InceptionV3 and DenseNet-121 from Pytorch \cite{paszke2019pytorch}.

\subsection{Experiments on Adversarial Perturbations in Frequency}
\label{eapf}
We analyze the distribution of adversarial perturbations in the frequency domain using a combination of three datasets (ImageNet-1K, CIFAR-10, and CIFAR-100), four models (VGG16, ResNet-50, DenseNet-121, and InceptionV3), and two attacks (FGSM and PGD). By computing the cosine similarity between adversarial and pure examples, we assess the distribution of adversarial perturbations across various frequency bands, as detailed in Table \ref{tab22}. The results reveal a consistent trend across diverse datasets, models, and attack strategies: the cosine similarity between normal and adversarial samples is markedly lower within $\{da\}$, $\{daa\}$ and $\{dad\}$ frequency bands. This trend suggests that adversarial perturbations predominantly occur in high-frequency bands of low-frequency bands. Moreover, although perturbations are distinctly observed in certain frequency bands, the complete separation of adversarial perturbations remains elusive, both through the use of different wavelet functions and further decomposition of the third-layer frequency band. This suggests that current wavelet functions are inadequate for the separation of adversarial perturbations. Designing new wavelet functions specifically for the characteristics of adversarial perturbations might facilitate a more complete separation of these perturbations.

\subsection{Evaluations of Our Proposed Method}
\label{sec6.3}
In this section, we evaluate our method on multiple datasets and models and also compare it with state-of-the-art approaches. All experiments are untargeted attacks. For performance evaluation and comparison, we employ the Attack Success Rate (ASR) and six other distinct metrics. These include the Average Number of Queries (ANQ), Median Number of Queries (MNQ), $L_2$ norm, $L_{\infty}$ norm, Structural Similarity Index (SSIM), and our proposed NDV.

\textbf{Evaluations on Different Models.} We evaluate our proposed attack method on DenseNet121, ResNet50 and InceptionV3 using CIFAR-10 and CIFAR-100 datasets. As shown in Table \ref{tab3}, the results demonstrate that our method can efficiently attack various models, with ASR exceeding 98\%. 
In particular, for the InceptionV3 model, an average of only a few dozen queries is needed for a successful attack.


\begin{table}[h]
	\centering
	
	\scalebox{0.73}{
		\begin{tabular}{llrrrrrrr} 
			\toprule
			\multicolumn{1}{c}{}
			& \multicolumn{1}{c}{Models} & ASR$\uparrow$ & ANQ$\downarrow$   & MNQ$\downarrow$ & $L_2$$\downarrow$ & $L_{\infty}$$\downarrow$ & SSIM$\uparrow$ & NDV$\downarrow$~  \\ 
			\toprule
			\multirow{3}{*}{C10}  & Den    & 0.948    & 219.63 & 176    & 2.27    & 0.18                           & 0.99   & 0.744              \\
			& R50    & 0.995 & 235.07 & 180 & 2.35 & 0.19 & 0.99 & 0.760             \\
			& V3     & 1.000        & 80.14   & 54     & 1.23    & 0.13                            & 0.99   & 0.403               \\ 
			\toprule
			\multirow{3}{*}{C100} & Den    & 0.997    & 179.49      & 137    & 2.06     & 0.18                            & 0.99   & 0.676              \\
			& R50    & 0.998    & 216.28 & 154    & 2.24    & 0.19                            & 0.99   & 0.736              \\
			& V3     & 0.995    & 59.22 & 42     & 1.12    & 0.12                              & 0.99   & 0.372              \\
			\bottomrule
			\multicolumn{1}{c}{}
			
		\end{tabular}
	}
	\caption{Evaluations on Different Models. Target models are DenseNet-121 (Den), ResNet-50 (R50) and InceptionV3 (V3). Datasets are CIFAR-10 (C10) and CIFAR-100 (C100). $\uparrow$ means the value is higher the better, and vice versa.}
	\label{tab3}
	\vspace{-0.1in}
\end{table}

\begin{table}[h]
	\centering
	
	\scalebox{0.68}{
		\begin{tabular}{llrrrrrrr}
			\toprule
			
			\multicolumn{1}{c}{}
			Methods                     & \multicolumn{1}{c}{Models} & ASR$\uparrow$ & ANQ$\downarrow$   & MNQ$\downarrow$ & $L_2$$\downarrow$ & $L_{\infty}$$\downarrow$ & SSIM$\uparrow$ & NDV$\downarrow$      \\
			\toprule
			\multirow{2}{*}{Boundary}   & Den    & 1.000   & 9077.00 & 5700  & 13.98 & 0.11 & 0.78 & 0.092  \\
			& R50    & 1.000     & 9820.06 & 7922  & 13.98 & 0.11 & 0.79 & 0.092  \\ \cline{1-9}
			\multirow{2}{*}{GeoDA}        & Den    & 1.000     & 3350.96 & 4078  & 32.99 & 0.35 & 0.87 & 0.223   \\
			& R50    & 1.000     & 3213.37 & 3329  & 30.98 & 0.33 & 0.87 & 0.210    \\ \cline{1-9}
			\multirow{2}{*}{SimBA} & Den    & 0.986 & 508.02 & 341   & 4.59 & 0.05 & 0.99 & 0.031  \\
			& R50    & 0.984 & 560.02 & 349   & 4.65 & 0.05 & 0.99 & 0.031   \\ \cline{1-9}
			\multirow{2}{*}{Square}  & Den    & 0.888 & 365.91 & 86    & 13.66 & 0.71 & 0.94 & 0.153   \\
			& R50    & 0.886 & 429.49 & 71    & 13.64 & 0.71 & 0.94 & 0.155   \\ \cline{1-9}
			\multirow{2}{*}{$\mathcal{N}$ Attack}  & Den    & 0.879 & 834.93 & 613    & 10.84 & 0.05 & 0.94 & 0.072   \\
			& R50    & 0.927 & 663.39 & 409    & 11.39 & 0.05 & 0.94 & 0.076   \\ \cline{1-9}
			\multirow{2}{*}{\textbf{Ours}}      & \textbf{Den}    &\textbf{0.995} & \textbf{380.58} & \textbf{183}   & \textbf{12.53} &\textbf{0.29} & \textbf{0.96} & \textbf{0.084}  \\ 
			&\textbf{R50}    & \textbf{0.992} & \textbf{426.11} &\textbf{184} & \textbf{12.85}  & \textbf{0.29} & \textbf{0.96} & \textbf{0.086} \\
			\bottomrule
			\multicolumn{1}{c}{}
			
		\end{tabular}
	}
	\caption{Comparisons with different methods on ImageNet-1K. Target models are DenseNet-121 (Den), ResNet-50 (R50).}
	\label{tab4}
	\vspace{-0.1in}
\end{table}

\begin{figure}[h]
	\centering
	\begin{subfigure}{0.49\linewidth}
		\includegraphics[width=1\linewidth]{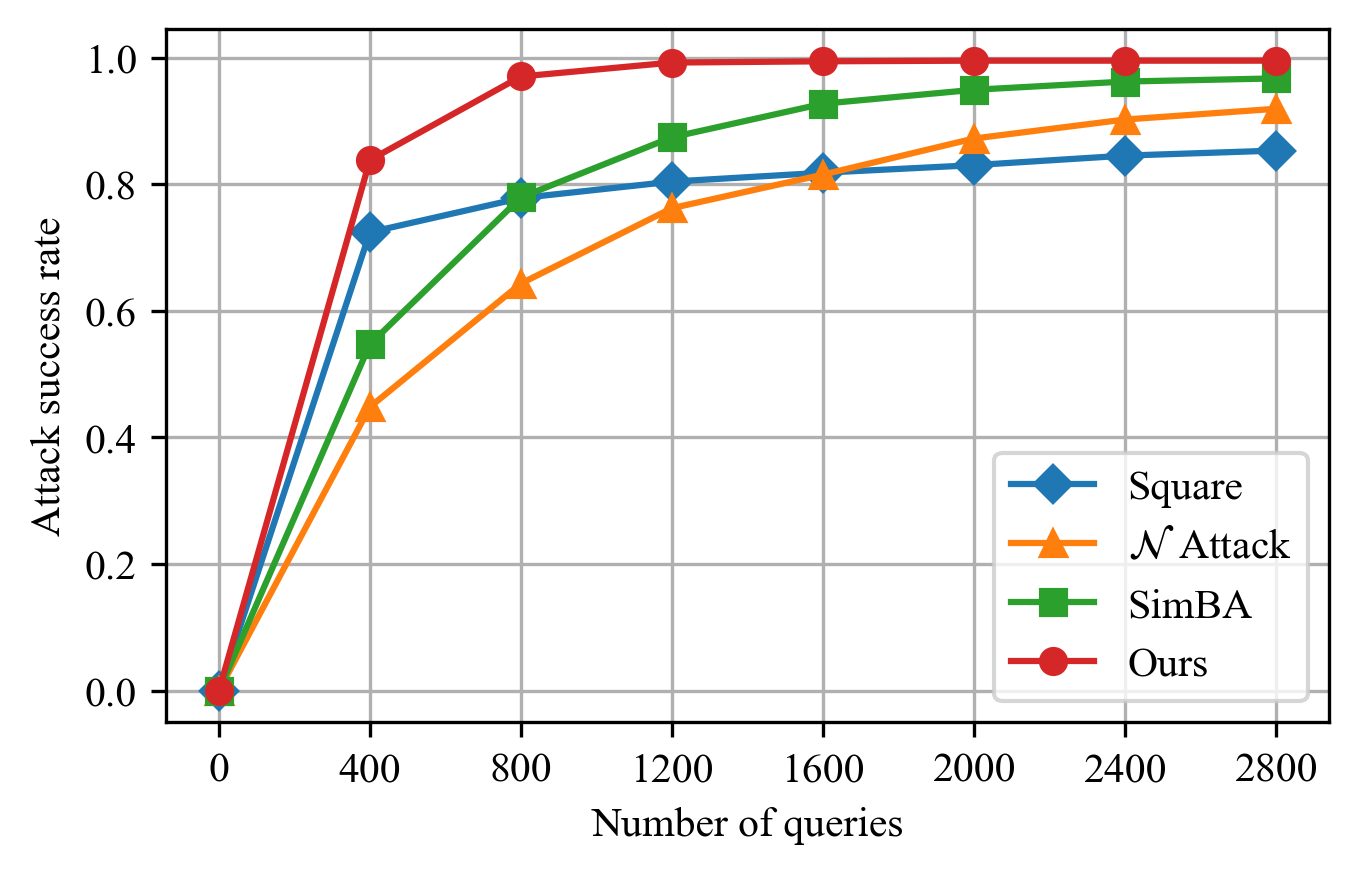}
		
	\end{subfigure}
	\hfill
	\begin{subfigure}{0.49\linewidth}
		\includegraphics[width=1\linewidth]{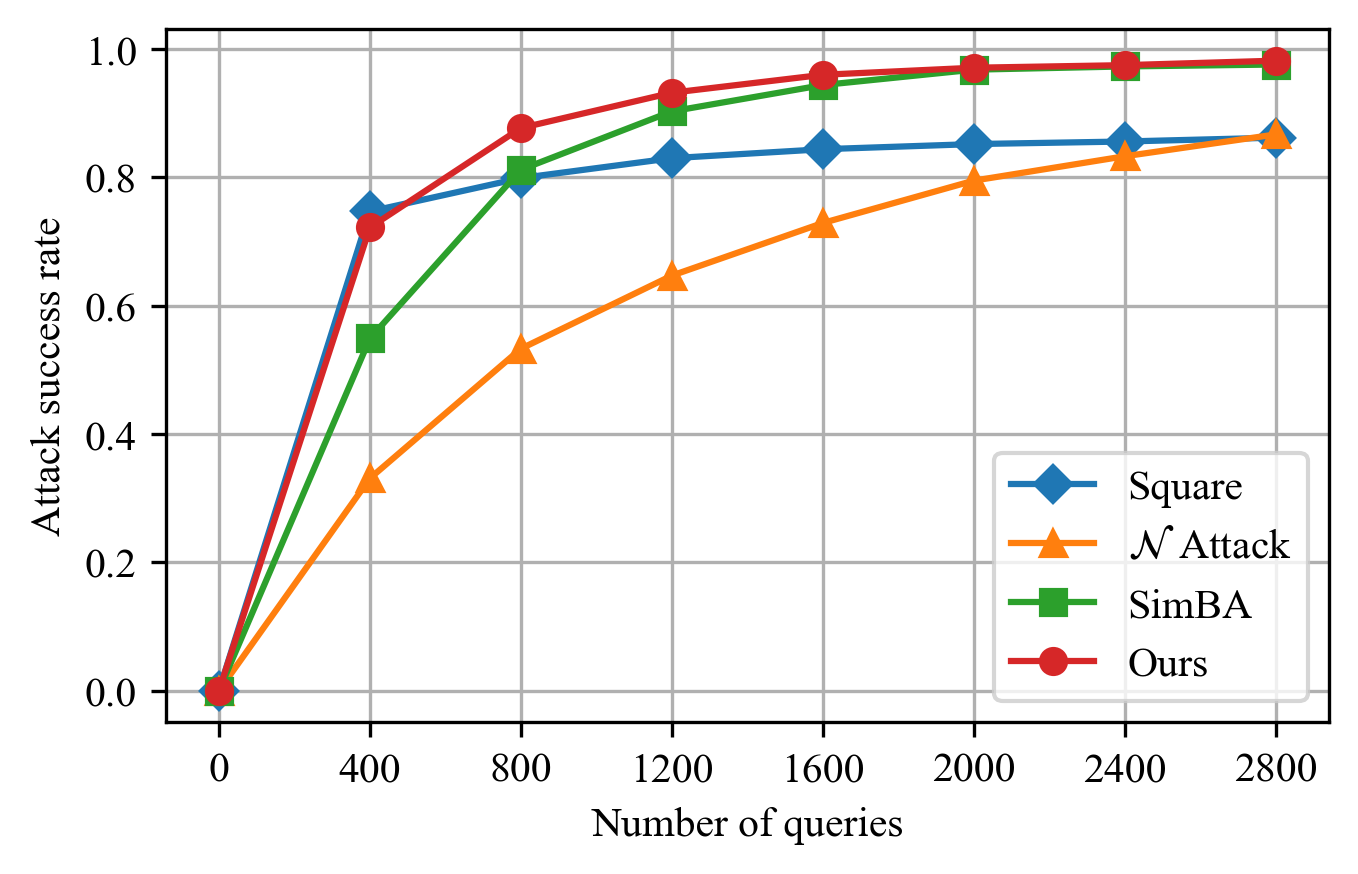}
		
	\end{subfigure}
	\caption{Performance of various attacks on ResNet-50 (left) and DenseNet-121 (right).}
	\label{fig7}
	\vspace{-0.1in}
\end{figure}

\textbf{Comparisons with State-of-the-Art Methods.} We compare our attack with state-of-the-art black-box attack algorithms, including Boundary Attack \cite{brendel2017decision}, GeoDA attack \cite{rahmati2020geoda}, Square Attack \cite{andriushchenko2020square}, $\mathcal{N}$ Attack \cite{li2019nattack}, SimBA \cite{guo2019simple}. As shown in Table \ref{tab4}, our attack achieves a fine balance between attack success, imperceptibility, and query efficiency, marking it as a superior strategy for adversarial attacks in practical applications. Figure \ref{fig7} illustrates the relationship between the number of queries and attack success rate, with results indicating that our method can achieve a high success rate with fewer queries.

\label{ab}
\textbf{Ablation Studies.} We investigate the contributions of three selected frequency bands—denoted as A ($\{aaa\}$), B ($\{daa\}$), and C ($\{dad\}$)—in generating adversarial perturbations, with the results detailed in Table \ref{tab2}. Band A demonstrates the highest individual ASR, underscoring its significant role in adversarial attacks, aligning with the studies of Wang et al \cite{wang2020towards}.  Combinations AB and AC yield higher ASR than each band individually, suggesting that the integration of bands B and C enhances attack efficacy. Notably, the ABC combination achieves the highest ASR, outperforming all single and dual-band combinations, while other evaluated metrics remain relatively unchanged.

\begin{table}[h]
	\centering
	
	\scalebox{0.85}{
		\begin{tabular}{l|rrrrrrr}
			\toprule
			\multicolumn{1}{c}{}
			& ASR$\uparrow$ & ANQ$\downarrow$   & MNQ$\downarrow$ & $L_2$$\downarrow$ & $L_{\infty}$$\downarrow$ & SSIM$\uparrow$ & NDV$\downarrow$  \\ 
			\toprule
			A   & 0.955 & 181.74 & 153 & 1.79 & 0.12 & 0.99 & 0.59 \\ 
			B   & 0.637 & 263.20 & 251 & 2.20 & 0.14 & 0.99 & 0.72 \\
			C   & 0.822 & 219.00 & 192 & 1.94 & 0.14 & 0.99 & 0.63 \\ \cline{1-8}
			AB  & 0.982 & 271.47 & 216 & 2.66 & 0.21 & 0.99 & 0.87 \\
			AC  & 0.987 & 216.16 & 166 & 2.27 & 0.17 & 0.99 & 0.74 \\
			BC  & 0.907 & 264.38 & 198 & 2.42 & 0.20 & 0.99 & 0.79 \\ \cline{1-8}
			ABC & 0.995 & 235.07 & 180 & 2.35 & 0.19 & 0.99 & 0.76 \\
			\bottomrule
		\end{tabular}
		
	}
	\caption{The influence of different combinations of three frequency bands on experimental results. We use A for $\{aaa\}$, B for $\{daa\}$, and C for $\{dad\}$.}
	\label{tab2}
	\vspace{-0.1in}
\end{table}

\subsection{Comparisons of $L_{2}$ and NDV}

\begin{figure}[h]
	\centering
	\includegraphics[width=1\linewidth]{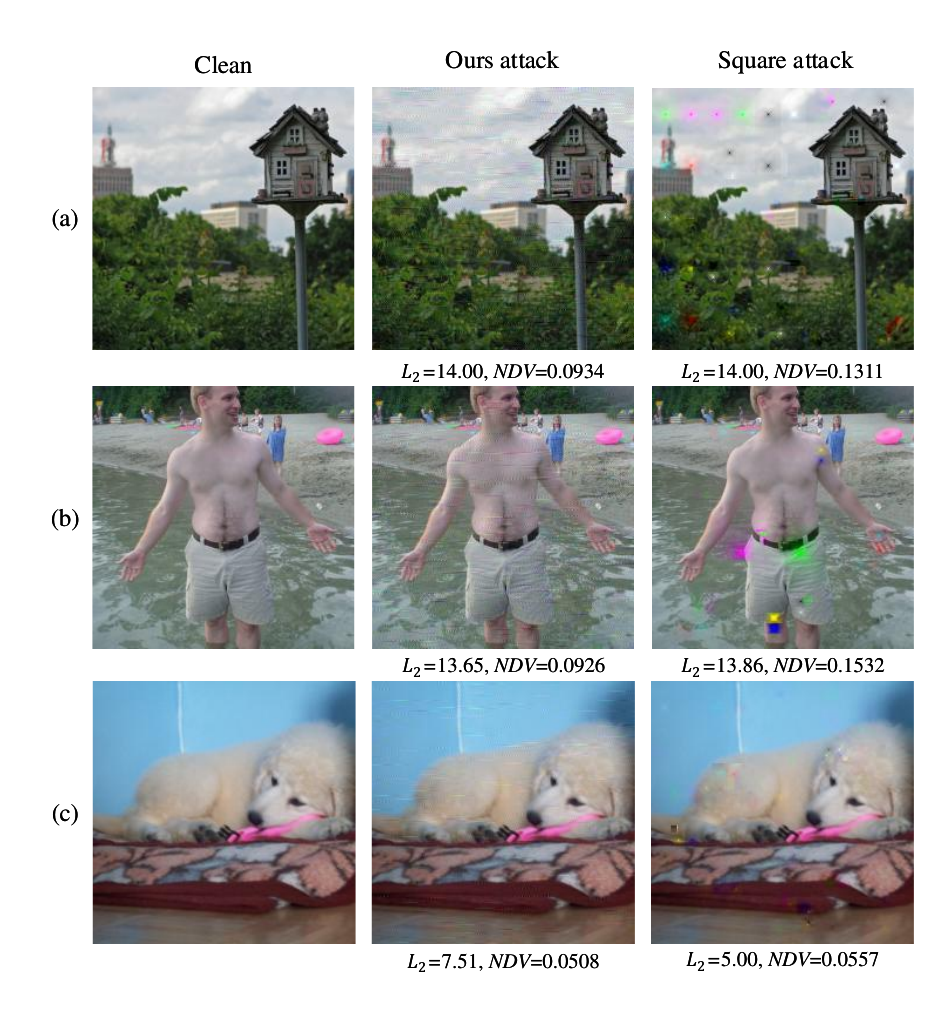}
	\caption{Visualizations of $L_2$ and NDV.}
	\label{fig5}
	\vspace{-0.1in}
\end{figure}

To enhance the comparative analysis between NDV and the $L_2$ norm, we select three sets of examples, as demonstrated in Figure \ref{fig5}. The first column displays the clean samples, while the second and third columns showcase adversarial examples created by our proposed method and the square attack method, respectively. The outcomes reveal that, despite nearly identical $L_2$ norms, the adversarial examples produced by the square attack exhibit more noticeable color variations and artifacts, thus making them more discernible to the human eye. In contrast, the changes in NDV values more accurately reflect the perceptual sensitivity of human observers.

\section{Conclusion}
Our investigation into the frequency attributes of adversarial perturbations has provided new insights into the comprehension of adversarial examples. The result of wavelet packet decomposition reveals that most perturbations are present within the lower-frequency components of high-frequency bands, thus refuting the simplistic dichotomy of perturbations as merely low or high-frequency.  Based on these insights, we propose a black-box adversarial attack algorithm that strategically employs combinations of different frequency bands to enhance attack efficacy. This innovative approach not only broadens the understanding of frequency-based adversarial strategies but also demonstrates higher efficiency compared to attacks that do not consider specific frequency components. Furthermore, to address the inadequacies of the $L_2$ norm for evaluating perturbations, we propose the NDV, which provides a more comprehensive measure that aligns closely with human visual perception. Our findings emphasize the value of a frequency-centric perspective in developing secure machine learning frameworks and advancing defenses against adversarial examples.
\bibliographystyle{named}
\bibliography{ijcai24}

\end{document}